\documentclass[letterpaper, 10 pt, conference]{ieeeconf}

\IEEEoverridecommandlockouts
\title{\LARGE \bf
Robots Ask the Way: Communication-Enabled Social Navigation
}

\author{Valentino Sacco* and Luca Scofano* and Indro Spinelli and Fabio Galasso
\thanks{*Authors contributed equally.}
\thanks{Sapienza University of Rome, Italy, email: lastname@di.uniroma1.it}}

\newcommand{\taskname}{CommNav}
\newcommand{\modelname}{COMM}
\newcommand{\acknowledgment}[1][]{\par%
    \phantomsection%
    \@ifmtarg{#1}{%
        \section*{Acknowledgment}%
        \addcontentsline{toc}{section}{Acknowledgment}%
    }{%
        \section*{Acknowledgment \\* #1}%
        \addcontentsline{toc}{section}{Acknowledgment: #1}%
    }%
}

\usepackage[most]{tcolorbox}
\usepackage[normalem]{ulem}
\usepackage{subcaption}
\usepackage{booktabs}
\usepackage{graphicx}
\definecolor{cvprblue}{rgb}{0.21,0.49,0.74}
\usepackage[pagebackref,breaklinks,colorlinks,allcolors=cvprblue]{hyperref}

\usepackage[dvipsnames]{xcolor}
\usepackage{amsmath}
\usepackage{amssymb}
\usepackage{svg}
\usepackage{float}
\usepackage{censor}

\begin{document}

\maketitle
\thispagestyle{empty}
\pagestyle{empty}

\begin{abstract}
Assistive autonomous robots operating in multi-agent environments require efficient strategies to locate specific individuals among multiple residents. Current social navigation methods focus on reactive collision avoidance and trajectory adaptation, but lack mechanisms to proactively gather information through human-robot communication.

We introduce Communication-enabled Social Navigation (CommNav). In this novel task, robotic agents actively seek assistance from residents to locate target individuals by requesting information about recent sightings, locations, and movements.

To evaluate CommNav, we extend Habitat 3.0 to create Habitat 3.0c, a communication-enabled variant supporting multi-human environments with information exchange protocols. Adding our communication module (COMM) to a state-of-the-art social navigation model yields a 10 percentage-point improvement in Episode Success. We further investigate the transition from structured data to natural language by evaluating models trained on LLM-generated instructions and on colloquial instructions collected from a human study.

Our experiments reveal that: (i) explicit human-robot communication substantially enhances multi-person navigation performance; (ii) pre-training COMM on a communication pretext task effectively addresses the challenge of occasional interaction signals; and (iii) the navigation policy is highly robust to natural, colloquial human language, achieving an episode success statistically similar to the model using perfect structured data.
\end{abstract}

\begin{figure*}[!t]
    \centering
    \includegraphics[width=400pt]{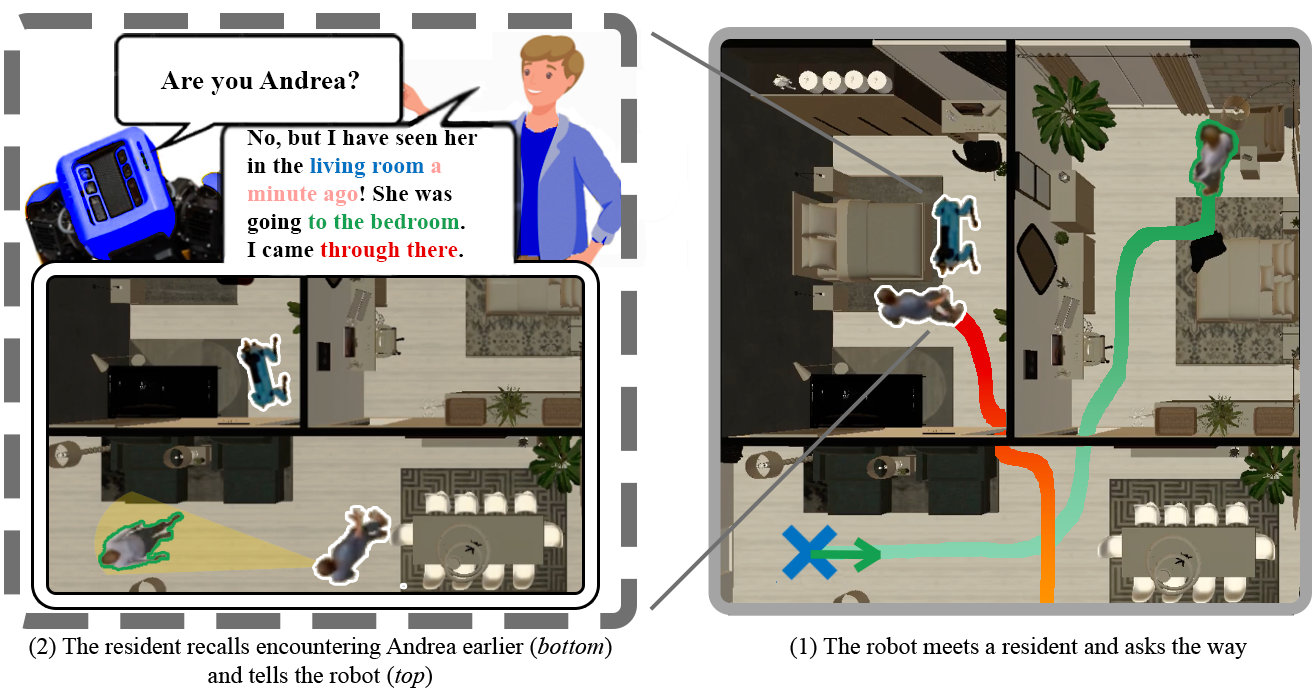}
\caption{In \textbf{CommNav}, the robot asks residents for help. If the resident is not Andrea, they may still provide cues---whether they \texttt{$\text{have seen}$} ($x_h$) her, \textcolor{Salmon}{\texttt{$\text{when}$}} ($x_t$), her \textcolor{NavyBlue}{\texttt{$\text{location}$}} ($\mathbf{x}_l$) and \textcolor{ForestGreen}{\texttt{direction}} ($\mathbf{x}_d$), and their own \textcolor{BrickRed}{\texttt{path}} ($\mathbf{x}_p$)---encoded as input to the navigation policy (Sec.~\ref{subsec:comm_message_structure}).}
\label{fig:teaser_image}
\end{figure*}

\section{Introduction}
\label{sec:intro}

Social navigation requires robots to avoid obstacles, interpret human behavior, and adapt to dynamic, partially unknown environments. Existing approaches have primarily emphasized collision avoidance and motion planning~\cite{Che2018EfficientAT}, yet they lack proactive mechanisms for acquiring task-relevant information from other agents. This limitation becomes critical when robots must identify and interact with specific individuals in shared spaces such as households, offices, or care facilities. For instance, a robotic assistant tasked with delivering medication would, under traditional methods, rely on exhaustive room-by-room searches. In contrast, humans naturally employ communication to obtain spatial cues~\cite{pramanick2024much,daniel1998spatial,senintegration}, and prior work confirms that dialogue improves cooperative navigation~\cite{amoozandeh2024,rehrl2009analysis} and that intermittent language instructions can boost long-horizon task performance~\cite{shi2024yellrobotimprovingonthefly}. State-of-the-art social navigation methods~\cite{Puig23,Scofano24,10801513} cannot leverage human knowledge of recent spatial events or exploit the collaborative potential of human-robot cohabitation, and incorporating communication introduces challenges related to sparse interaction signals, heterogeneous cue types, and scalability as the number of agents grows.

We introduce \textbf{Communication-enabled Social Navigation (\taskname)}, a novel task where robotic agents actively seek assistance from residents to locate target individuals, as illustrated in Fig.~\ref{fig:teaser_image}. We formalize CommNav as a multi-agent navigation problem where robots query human agents for spatio-temporal information about a target's location. Instead of exhaustive search, the robot can approach a resident and ask, ``Have you seen them?'' A helpful response, such as ``I saw them entering the kitchen a few minutes ago,'' allows the robot to navigate far more efficiently.\footnote{This work considers various degrees of human accuracy, e.g.\ the cue may also read: ``I saw them 7 meters North, heading left, a minute ago.'' See Sec.~\ref{subsec:comm_message_structure} for details.}

To enable this skill, we introduce a novel \textbf{\modelname~module} integrated into a leading DDPPO navigation model. Learning from the sparse and unpredictable signals of human interaction is a significant challenge. We address this by pre-training \modelname~on a communication pretext task, which achieves significant performance improvements over non-communicative baselines. A key part of our investigation is bridging the gap between structured data (like coordinates or timestamps) and natural human interaction. We test our \modelname~architecture with three data types: (1) perfect structured data, (2) synthetically precise instructions generated by an LLM (\modelname$_{\mathcal{L}}$), and (3) natural, colloquial instructions from a human study (\modelname~$_{\mathcal{L}(Human)}$). Our findings reveal a surprising insight: while human inputs are often underspecified compared to LLM precision, localization error and model performance are statistically indistinguishable from their synthetic counterpart. Code can be found at: \url{https://github.com/S4b3/CommNav}.

\section{Related Work}
\subsection{Embodied Navigation}
Recent progress in embodied navigation has been fueled by 3D indoor datasets~\cite{chang2017matterport, shen2021igibson, ramakrishnan2021habitat} and advanced simulators~\cite{savva2019habitat, shen2021igibson, kolve2017ai2}. Most conventional simulators represent humans as static objects or basic obstacles~\cite{xia2020interactive, yokoyama2021learning}, limiting research that requires dynamic human-robot interactions. More recent platforms, including Habitat 3.0~\cite{Puig23}, PARTNR~\cite{chang2024partnr}, and others~\cite{Li2024CoNavAB, An23}, support complex human dynamics, facilitating socially aware studies. PARTNR~\cite{chang2024partnr} introduces a semi-automated task-generation pipeline for human-robot interaction reasoning tasks, leveraging LLMs to construct richer evaluation scenarios. Beyond navigation, recent work explores multi-agent cooperation with language-enabled agents in task-oriented environments~\cite{zhang2024building,liu2025capo}. Our \taskname~goes further by incorporating active human-robot collaboration in dynamic, mapless, multi-human social settings.

\subsection{Social Navigation and Dynamic Environments}
Simulated environments such as iGibson~\cite{Li2021iGibson2O}, SEAN~\cite{Tsoi2022SEAN2F}, VirtualHome~\cite{Puig2018VirtualHomeSH}, and Habitat 3.0~\cite{Puig23} provide privileged information about scene and agent elements, which is valuable for training RL policies.
While leveraging such information to forecast human trajectories can enhance navigation~\cite{Patle2019ARO}, communication-based cues are inherently sparse and intermittent, requiring the system to cope with limited social information. To address this, we draw inspiration from asynchronous state-action history identification methods~\cite{kumar2021rma, loquercio2023learning, zhang2023, Scofano24}: unlike prior work that infers state-action mappings in a separate training stage, we integrate a pre-trained module directly into an online policy, enabling it to utilize sparse social cues effectively.

Research on socially-aware navigation aims to enable robots to navigate shared spaces while respecting human social norms: classical collision-avoidance methods~\cite{Berg11,rvo,cadrl} have been extended with social rules and proxemic constraints~\cite{sa-cadrl,socialforces,cancelli23}, while recent policies model crowd-level interactions through social attention and spatiotemporal graphs~\cite{socialattention,social-graph}. Communication has also been used as an implicit or explicit channel to make robot intent legible, and natural gestures have been explored to guide embodied agents~\cite{Che2018EfficientAT,Wu2021}; in contrast, \taskname{} studies information-seeking communication, where the robot actively queries non-target humans for sparse egocentric cues about another person and integrates them into mapless multi-human search.

\begin{figure*}[!t]
    \centering
    \includegraphics[width=0.75\linewidth]{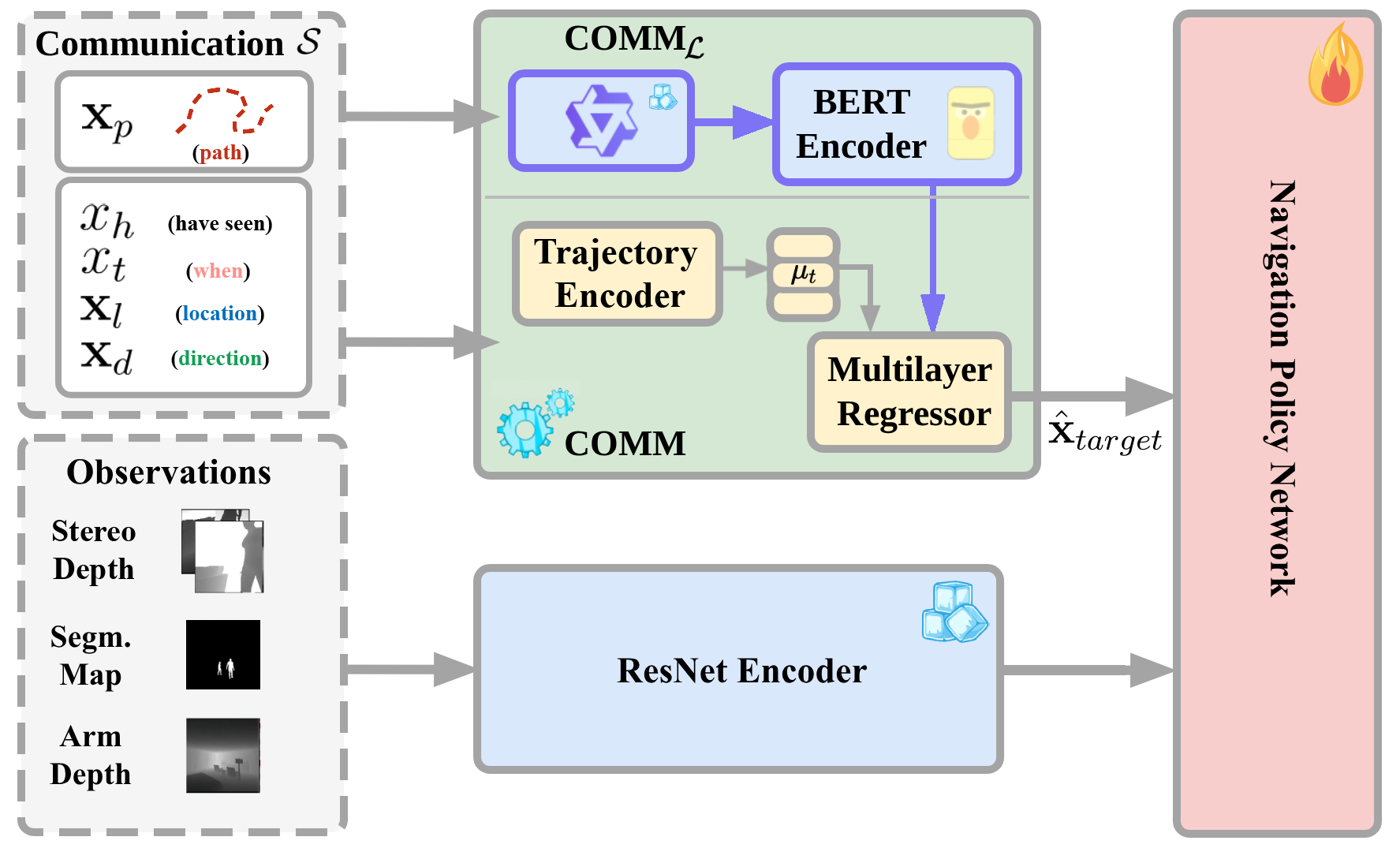}
    \caption{Architecture of the COMM model. In the structured path, $\mathbf{x}_p$ is first encoded as a trajectory embedding and then concatenated with $[x_h, x_t, \mathbf{x}_l, \mathbf{x}_d]$ to regress $\hat{\mathbf{x}}_{\text{target}}$. In the language path, the full state $\mathcal{S}$ is verbalized by QWEN3-8B into $\mathcal{L}$, which a frozen BERT encoder maps to the same target-location estimate; this prediction is fused with ResNet-encoded observations by the Navigation Policy.}
    \label{fig:model_architecture}
\end{figure*}

\section{Communication-enabled Social Navigation}

We introduce Communication-enabled Social Navigation (\taskname), where a robot actively seeks and follows human guidance in real-world environments. Unlike traditional social navigation, which emphasizes collision avoidance and social norms, \taskname~leverages communication to improve efficiency and cooperation. By asking for and interpreting directions, robots can reduce exploration time and adapt their behavior in crowded or unfamiliar spaces~\cite{burgoon1995, Joo2017, Trabelsi2019, Doherty2022}.

In this setting, multiple humans may have observed the individual of interest and are willing to help. The robot does not target a specific person, but opportunistically collects cues from anyone it encounters. Once it locates the individual and receives consent, it begins providing assistance. \taskname~thus extends social navigation with communication-driven interaction, emphasizing adaptation in dynamic, multi-agent environments~\cite{Trabelsi2019, Wu2021, ajina2023, Liang19}.

\subsection{Content of Communication Messages}\label{subsec:comm_message_structure}

Formally, when the robot is close to and oriented toward a human (i.e., within its field of view), it receives a message encoded as a state vector $\mathcal{S}$ with the following components: 
\begin{itemize}
    \item $x_h$ (\texttt{have seen}), a boolean indicating whether the speaker has observed the target;
    \item $x_t$ (\textcolor{Salmon}{\texttt{when}}), a scalar counting steps since the target was last seen;
    \item $\mathbf{x}_l$ (\textcolor{NavyBlue}{\texttt{location}}), the last known 3D location of the target agent $\mathbf{x}_l\in\mathbb{R}^{3}$;
    \item $\mathbf{x}_d$ (\textcolor{ForestGreen}{\texttt{direction}}), a 3D point $\mathbf{x}_d\in\mathbb{R}^{3}$ representing the direction of the target human;
    \item $\mathbf{x}_p$ (\textcolor{BrickRed}{\texttt{path}}), 100 3D point coordinates $\mathbf{x}_p\in\mathbb{R}^{100\times 3}$ consisting of the speaker's past trajectory.
\end{itemize}
We use non-bold $x$ for scalar or Boolean cues and bold $\mathbf{x}$ for vector-valued spatial cues. All coordinates are expressed relative to the speaking agent's current position, yielding an egocentric description of the encounter.

If the target agent is not in sight, the communication sensor updates only $\mathbf{x}_p$ and $x_t$, provided the target has been previously encountered. When the target agent is encountered, the human updates $\mathbf{x}_l$ and $\mathbf{x}_d$; at this point, $x_h$ becomes \textit{True}, and $x_t$ begins updating. If the agents meet again, all encounter-related variables are overwritten.

Whereas real human communication is approximate and often incomplete, 
$\mathcal{S}$ provides a precise and exhaustive representation of target encounters. This structured information serves as a foundation for generating natural language instructions. To close the gap with real-world scenarios, we use the QWEN3-8B model~\cite{yang2025qwen3technicalreport} to translate each state vector into a short utterance $\mathcal{L}$; this conversion is model-agnostic and any instruction-following LLM can be substituted.

The following example illustrates the outcome of this conversion into a more natural language exchange.

\textbf{Ex.~1:} \textit{Input $\mathcal{S}$:} \(x_h=1,\;x_t=53,\;\mathbf{x}_p\in\mathbb{R}^{100\times 3}\). \textit{Output $\mathcal{L}$:} ``I saw them about 4\,m in front of me and slightly to my right; they were heading left. I approached from behind.''; \quad

\textbf{Ex.~2:} \textit{Input $\mathcal{S}$:} \(x_h=0,\;\mathbf{x}_p\in\mathbb{R}^{100\times 3}\). \textit{Output $\mathcal{L}$:} ``No. I haven't seen them. I came from about 6\,m behind and from my left.''

Our instructions avoid place names (e.g., `kitchen') by design, as semantic grounding would shift our egocentric formulation toward object goal navigation. By restricting inputs to relative directions (e.g., ``behind me''), we ensure the policy learns spatial reasoning from social cues rather than relying on static semantic look-ups.

\taskname{} therefore evaluates not only whether communication improves navigation, but also how effectively structured representations can be grounded into a more natural interaction. Implementation details are provided in the Appendix.

\subsection{RL with Human-Robot Communication}

Our approach, illustrated in Figure~\ref{fig:model_architecture}, extends a standard ResNet-based~\cite{7780459} navigation architecture with a dedicated \textbf{Communication Module (COMM)} that provides a parallel stream of guidance. The baseline ResNet encoder processes raw observations (RGB-depth) into a compact representation. In parallel, the COMM module interprets intermittent human-provided inputs ($\mathcal{S}$) and produces an estimated target location, treated as an additional sensor. 

The \textbf{Navigation Policy Network} then fuses the two streams to generate the actions. When human input is unavailable, the COMM module substitutes a placeholder, and it allows the agent to fall back on its visual policy. This design enables occasional use of human guidance while maintaining complete autonomy.

\textbf{Policy Training.} As for state-of-the-art models~\cite{Puig23, Scofano24}, the navigation policy follows the DDPPO framework and is trained with reinforcement learning objectives tailored to \taskname, including success-based and efficiency-weighted rewards. Training begins without communication, ensuring the policy learns robust navigation from visual inputs alone.

\textbf{Communication Module.} The COMM module is pre-trained to interpret infrequent human-provided information. Rather than concatenating communication data with visual features, it translates cues into an estimated target position $\hat{\mathbf{x}}_{target}$, analogous to a PointGoal sensor~\cite{Zhao2021, Partsey2022}. 

We pre-train \modelname{} on the proxy task of predicting the target's current position from the interaction message. For this purpose we collect a dataset of $2.4\,$M communication instances of $\mathcal{S}$ gathered over $60$ million simulation steps of baseline training; each recorded communication instance contains $\mathcal{S}=\{x_h, x_t, \mathbf{x}_l, \mathbf{x}_d, \mathbf{x}_p\}$ together with the ground-truth target location $\mathbf{x}_{\text{target}}$, which serves as the regression target. Because the elements of $\mathcal{S}$ are heterogeneous, we process them separately and train \modelname{} in two complementary ways. For the input path, the trajectory \(\mathbf{x}_p\) is embedded by the Trajectory Encoder into a hidden vector \(\mu_t\in\mathbb{R}^h\). We concatenate \(\mu_t\) with the vector \([x_h, x_t, \mathbf{x}_l, \mathbf{x}_d]\) and feed the result to a Multilayer Regressor that outputs the estimated target location \(\hat{\mathbf{x}}_{\text{target}}\). Specifically, we employ a Spatio-Temporal MLP with $l{=}3$ attention heads and $H{=}64$ hidden dimensions for the Trajectory Encoder, and a 4-layer MLP for the Regressor.

Separately, to train a language-only variant, denoted \(\text{\modelname}_{\mathcal{L}}\), we convert a subset of the dataset into natural-language instructions: 7000 samples from the collected instances are processed with QWEN3-8B~\cite{yang2025qwen3technicalreport} to generate corresponding natural-language reports. \(\text{\modelname}_{\mathcal{L}}\) is then equipped with a BERT encoder \cite{devlin2019bertpretrainingdeepbidirectional} and trained to regress \(\mathbf{x}_{\text{target}}\) using these textual inputs alone. Both training regimes learn the same regression objective (predicting \(\mathbf{x}_{\text{target}}\)) but from different input modalities (numerical vs.\ natural language).

This explicit estimate enables the policy to incorporate communicative hints about the target's location and direction without requiring continuous updates, thereby tightly integrating human input with autonomous navigation.

\subsection{Habitat 3.0c}\label{sec:background_works}
We extend Habitat 3.0~\cite{Puig23}, a state-of-the-art Embodied AI simulator supporting tasks such as Social Rearrangement~\cite{Szot2023}, Social Navigation~\cite{Scofano24, campari2022online}, and PointGoal Navigation~\cite{Partsey2022, wijmans2020ddppolearningnearperfectpointgoal}, into a multi-human setting we call Habitat 3.0c. Here, the robot must identify and assist a target among several agents. Identification is triggered when the robot is within communication range and facing a human, defined as $\phi_T = \langle \mathbf{x}_{robot}, \mathbf{x}_{agent} \rangle$. The decision to approach a bystander is learned end-to-end by the navigation policy. If the individual is not the target, the robot receives an information set $\mathcal{S}$, which can help infer the target's location, analogous to a PointGoal sensor~\cite{Zhao2021, Partsey2022}. Since communication occurs only at encounters and is not guaranteed every episode, the robot must handle sparse and incomplete guidance.

To encourage realistic interactions, Habitat 3.0c integrates Optimal Reciprocal Collision Avoidance (ORCA)~\cite{Berg11} and introduces a probability $p=0.25$ that humans ignore the robot for an entire episode. This modification allows humanoid-initiated collisions, teaching the robot to yield dynamically. The robot's policy takes egocentric depth and a humanoid detector as input and outputs linear and angular velocity commands, making the approach robot-agnostic.

\section{Experiments}
\begin{table*}[!t]
\centering
\renewcommand{\arraystretch}{1.5} 
\setlength{\tabcolsep}{3pt}       
\caption{Dialogue-Driven Social Navigation Results. ``Int.'' specifies whether interaction between human and robot is enabled, while ``MH'' indicates the presence of multiple humans in the environment.}
\label{tab:main_table}

\resizebox{1\textwidth}{!}{%
  \begin{tabular}{l|cc|c|ccccccc}
  \hline
  Models & Int. & MH & Habitat & S ($\uparrow$) & $S_{\text{steps}}$ ($\downarrow$) & SPS ($\uparrow$) & F ($\uparrow$) & CR ($\downarrow$) & $\mathrm{CR}_T$ ($\downarrow$) & ES ($\uparrow$) \\
  \hline
  DDPPO~\cite{Puig23} & - & - & 3.0 
  & $0.76\pm0.02$ & $483\pm6.50$ & $0.34\pm0.01$ & $0.29\pm0.01$ & $0.48\pm0.03$ & - & $0.40\pm0.02$ \\
  SDA~\cite{Scofano24}  & - & - & 3.0 
  & $0.91\pm0.01$ & $415\pm8.00$ & $0.45\pm0.01$ & $0.39\pm0.01$ & $0.57\pm0.02$ & - & $0.43\pm0.02$ \\ 
  \hline\hline
  DDPPO~\cite{Puig23}  & - & \checkmark & 3.0c 
  & $0.71\pm0.02$ & $618\pm1.00$ & $\mathbf{0.40\pm0.02}$  & $0.10\pm0.01$ & $0.64\pm0.02$ & $0.30\pm0.01$  &  $0.14\pm0.02$\\
  SDA~\cite{Scofano24} & - & \checkmark & 3.0c 
  & $0.70\pm0.01$ & $613\pm3.50$ & $0.40\pm0.01$ & $0.10\pm0.01$ & $0.76\pm0.01$ & $0.27\pm0.02$ & $0.16\pm0.01$ \\
  DDPPO~\cite{Puig23}  & \checkmark & \checkmark & 3.0c 
  & $0.70\pm0.02$ & $608\pm4.00$ & $0.40\pm0.01$ & $0.10\pm0.01$ & $0.68\pm0.03$  & $0.30\pm0.02$ & $0.14\pm0.01$  \\
  \hline
  \modelname{} & \checkmark & \checkmark & 3.0c 
  & $\mathbf{0.78\pm0.01}$ & $572\pm2.00$ & $0.38\pm0.01$  & $\mathbf{0.13\pm0.01}$ & $\mathbf{0.51\pm0.02}$ & $\mathbf{0.23\pm0.02}$ & $\mathbf{0.24\pm0.02}$ \\ 
  $\text{\modelname}_{\mathcal{L}}$ & \checkmark & \checkmark & 3.0c & $\mathbf{0.78\pm0.01}$ & $\mathbf{546\pm5.00}$ & $0.39\pm0.01$ & $0.12\pm0.02$  & $0.59\pm0.01$ & $0.28\pm0.02$ & $0.20\pm0.01$  \\
  \hline
  $\text{\modelname}_{\mathcal{L}(Human)}$ & \checkmark & \checkmark & 3.0c & $0.72\pm{0.05}$ & $620\pm{60.00}$ & $0.38\pm{0.03}$ & $0.13\pm{0.01}$  & $0.51\pm{0.01}$ & $0.24\pm{0.02}$ & $0.23\pm{0.01}$  \\
  \hline
  \end{tabular}%
}
\end{table*}

\begin{figure*}[t]
    \centering
    \includegraphics[width=\linewidth]{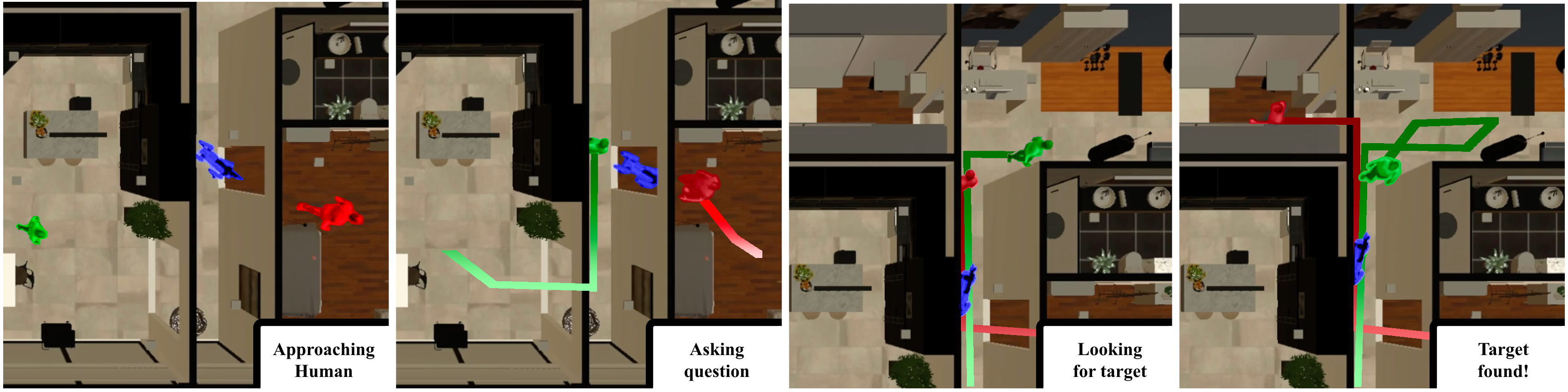}
    \caption{Sequence of human-robot communication for target localization. The \textcolor{MidnightBlue}{robot} approaches a \textcolor{BrickRed}{non-target human} (frame 1), who provides cues---\textcolor{Salmon}{when}, \textcolor{NavyBlue}{where}, the \textcolor{ForestGreen}{direction}, and their own \textcolor{BrickRed}{path} (frame 2). The \textcolor{MidnightBlue}{robot} adjusts its course (frame 3) and locates the \textcolor{ForestGreen}{target human} (frame 4).}
    \label{fig:qualitative_communication_slider}
\end{figure*}
\textbf{Baselines.} We evaluate two top-performing social navigation methods: DDPPO~\cite{Puig23}, a state-of-the-art RL approach using a recurrent neural network to generate actions from egocentric sensory inputs (stereo depth, human detector, GPS coordinates); and SDA~\cite{Scofano24}, which enhances real-time adaptation to human movement by encoding fully visible human trajectories as social dynamics cues. In Habitat 3.0c, we use SDA's first stage only, providing an upper bound by leveraging complete trajectory information. Evaluating these baselines without communication establishes how far current exploration strategies can reach in Habitat 3.0c before any communicative guidance is introduced.

\textbf{Metrics.} \textit{Finding Success (S)}: ratio of episodes where the agent located and reached the human; \textit{$\text{S}_{steps}$}: average steps to first find the human; \textit{SPS}: path efficiency relative to the optimal path; \textit{Following Rate (F)}: ratio of steps maintaining 1--2\,m distance while facing the human; \textit{CR}: ratio of episodes ending in any collision; \textit{$\text{CR}_T$}: ratio ending in collision with the target; \textit{Episode Success (ES)}: ratio of episodes where the agent found and followed the human for the required steps at a safe 1--2\,m distance. $\text{CR}_T$ is newly added to evaluate selective navigation around the target in multi-agent settings.

\subsection{\taskname\ Results}
\label{subsec:quant_results}

Table~\ref{tab:main_table} compares Social Navigation on Habitat~3.0 (top) with Communication-enabled Social Navigation on the extended Habitat~3.0c environment (bottom). Comparing reproduced single-human results (top)~\cite{Scofano24} with multi-human results (bottom) highlights how task difficulty increases when multiple residents are present. To adapt SDA~\cite{Scofano24} for the multi-human \taskname~task, we enhanced the model to capture the inherent social dynamics. In line with state-of-the-art, each evaluation consists of 400 test episodes, spread across 12 unseen test environments~\cite{Puig23, Scofano24}.

Transitioning from a single-human to a multi-human environment causes a significant drop in ES: from 0.40/0.43 to 0.14/0.16 for DDPPO/SDA. Training DDPPO with communication produces no gains (ES remains 0.14), showing that incorporating communication is non-trivial. \modelname~achieves S of 0.78 and ES of 0.24, improving over DDPPO w/ Interaction by 8 pp in Finding Success and 10 pp in Episode Success, demonstrating that proactive communication substantially improves the robot's ability to locate the target. In efficiency, \modelname~takes 572 $\text{S}_{\text{steps}}$ vs.\ 608--618 for baselines. Notably, $S_{\text{steps}}$ inherently penalizes query time; COMM's stable $S_{\text{steps}}$ confirms that communication improves search efficiency beyond query overhead, and the policy learns to query selectively.

Although SPS decreases slightly (0.40 to 0.38), reflecting detour steps taken to approach and query bystanders, this trade-off yields improved Finding Success (0.70 to 0.78). \modelname~achieves CR of 0.51, lower than 0.68 for DDPPO w/ Interaction and 0.64/0.76 for DDPPO/SDA: the communication task forces frequent close proximity with non-target humans, providing dense interaction-based supervision for avoidance, explaining why \modelname~achieves lower CR than baselines that avoid social contact entirely (see supplementary video). $\text{CR}_T$ is minimized to 0.23, underscoring safe target following.
Transitioning to generated language instructions, \modelname$_\mathcal{L}$ achieves Finding Success of $0.78$ with $S_{\text{steps}}{=}546$, closely matching \modelname{}. Episode-level metrics show modest degradation: ES drops from $0.24$ to $0.20$, CR increases ($0.51 {\rightarrow} 0.59$; $\mathrm{CR}_T$: $0.23 {\rightarrow} 0.28$), and Following decreases slightly ($0.13 {\rightarrow} 0.12$). While the model is largely robust across modalities, the added noise and underspecification of generated language instructions reduces episode-level reliability.

Overall, \modelname~sets a strong baseline for communication-enabled social navigation within complex, multi-human environments.
Qualitatively, Figure~\ref{fig:qualitative_communication_slider} shows how real-time human guidance helps the robot locate the target.

\subsection{Natural-language communication and human study}
\label{sec:nl_communication}
To study the gap between model-generated instructions $\mathcal{L}$ and human-provided directions, we ran a human study with $20$ participants: each participant was shown 50 visual representations of $\mathcal{S}$, out of the 7000 QWEN-converted samples, and asked to give directions to the robot. The study reveals that human responses often approximate the task but omit precise spatial or temporal details. For example, participants said \textit{``We were together 2 seconds ago, now he moved towards my right''} or \textit{``I met him 10 seconds ago''}, rather than specifying exact distances or orientations. In contrast, QWEN3-8B outputs explicitly encode fine-grained details, such as \textit{``I saw them last about 3 meters in front of me and a little to the left; they were headed forward and to the right. I came from about 3 meters behind me''}.

Overall, human guidance was simpler and less detailed than the task-specific instructions generated in \taskname{}$_\mathcal{L}$. To quantify this gap, we compared LLM-generated instructions with those produced by humans. At the sentence level, cosine similarity of the embeddings reaches $0.83$, indicating strong semantic alignment. At the token level, BERTScore (F1)~\cite{Zhang2019BERTScoreET} drops to $0.50$, highlighting reduced overlap in lexical and structural choices.

We test \(\text{COMM}_{\mathcal{L}}\) robustness to this \textit{style} of instruction, denoted \(\text{COMM}_{\mathcal{L}(Human)}\), by conducting a 100-episode evaluation.
For this experiment, simulation stopped whenever the robot initiated communication, in order for the participant to provide colloquial instructions (e.g., \textit{``No, I haven't seen them''}, \textit{``Yes, I saw them behind me. They were going left''}).

\begin{figure}[t]
    \centering
    \begin{subfigure}{0.49\linewidth}
        \centering
        \includegraphics[width=\linewidth]{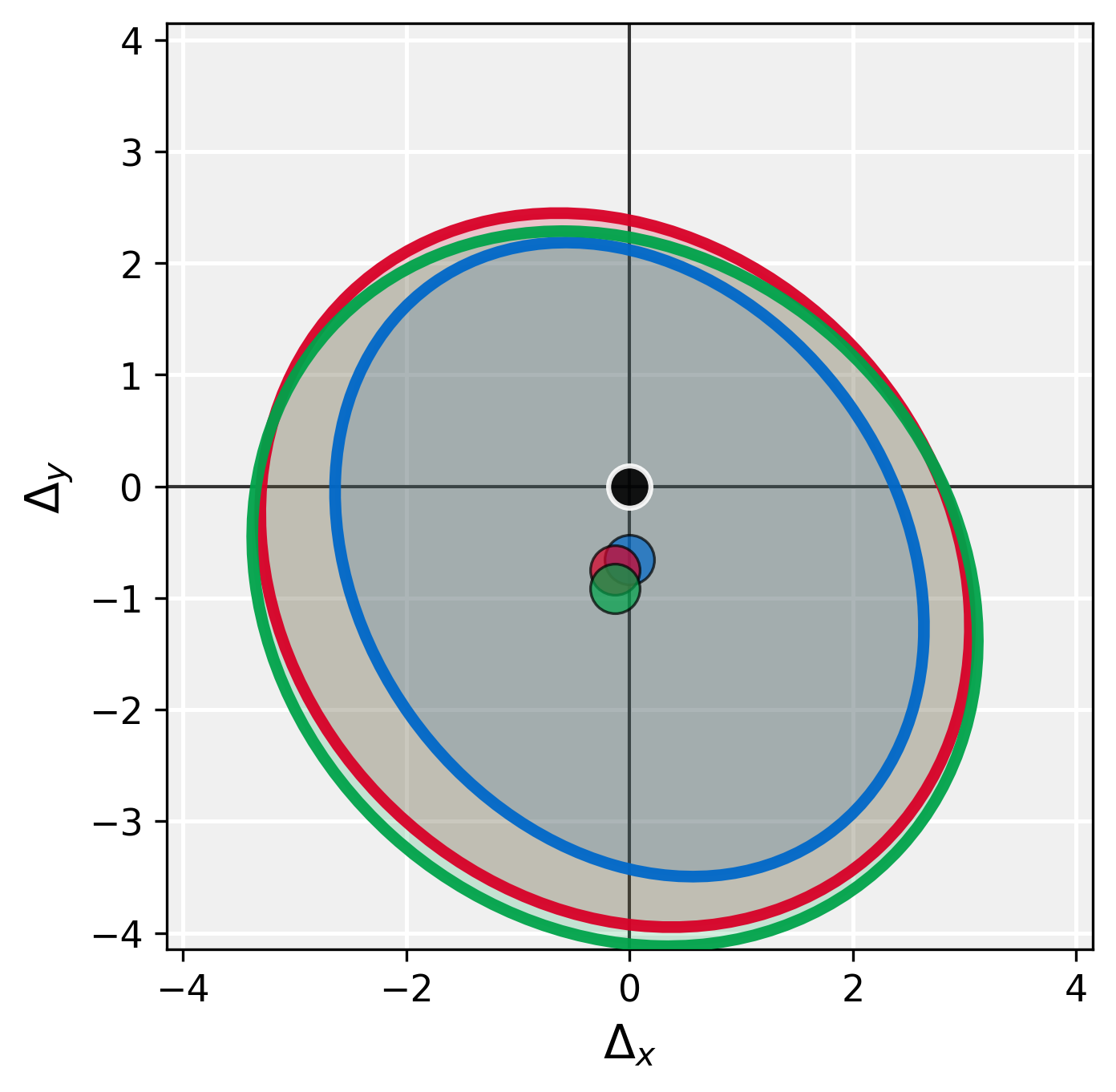}
        \caption{$x_h = 0$}
        \label{fig:ellipses_xh0}
    \end{subfigure}
    \hfill
    \begin{subfigure}{0.49\linewidth}
        \centering
        \includegraphics[width=\linewidth]{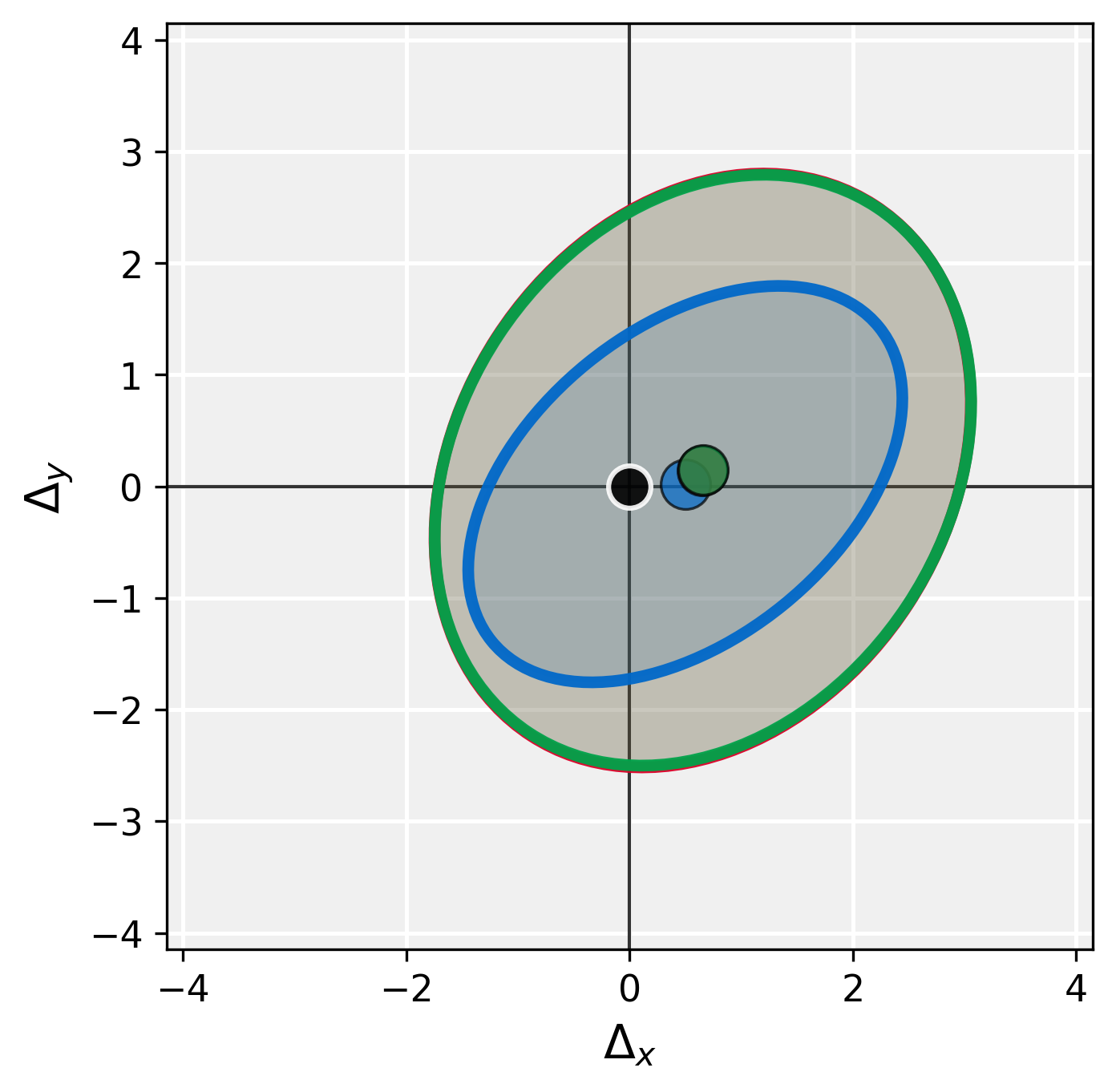}
        \caption{$x_h = 1$}
        \label{fig:ellipses_xh1}
    \end{subfigure}
    \caption{COMM target-prediction error in the ground plane. Each coloured dot is the mean error vector; the surrounding ellipse covers the $1\sigma$ spread. Closer to the origin ({\color{black}$\bullet$}) with a smaller ellipse indicates better localization. {\color[HTML]{0068c9}\textbf{Blue}}: \modelname; {\color[HTML]{d80027}\textbf{Red}}: \modelname$_{\mathcal{L}}$; {\color[HTML]{00a34a}\textbf{Green}}: \modelname$_{\mathcal{L}(\mathrm{H})}$.}
    \label{fig:ellipses_xh}
\end{figure}

As observed in Table~\ref{tab:main_table}, using colloquial human inputs, \(\text{COMM}_{\mathcal{L}(Human)}\) obtains larger standard deviation on Finding Success ($S$) than \(\text{COMM}_{\mathcal{L}}\), $0.72\pm{0.05}$ to~\(\text{COMM}_{\mathcal{L}}\)'s $0.78\pm{0.01}$, highlighting how less structured communication may affect guidance. \(\text{COMM}_{\mathcal{L}(Human)}\) achieves an Episode Success (ES) of $0.23 \pm 0.01$. This result is higher than the $0.20 \pm 0.01$ ES achieved by the LLM-generated \(\text{COMM}_{\mathcal{L}}\) model (Table~\ref{tab:main_table}). Furthermore, \modelname$_{\mathcal{L}(Human)}$ achieves lower $\mathrm{CR}_T$ than \modelname$_{\mathcal{L}}$. To investigate, we plot the per-episode prediction error $e_i{=}\hat{\mathbf{x}}_{\text{target},i}{-}\mathbf{x}_{\text{target},i}$ in the ground plane (Fig.~\ref{fig:ellipses_xh}). While unseen episodes ($x_h{=}0$) show higher dispersion than seen ($x_h{=}1$), \modelname$_{\mathcal{L}}$ and \modelname$_{\mathcal{L}(\mathrm{Human})}$ ellipses are nearly identical within each split, with mean distances $\approx 2.52$ ($x_h{=}1$) and $\approx 3.8$ ($x_h{=}0$). This suggests the performance drop in~\modelname$_{\mathcal{L}}$ is not driven by localization outliers or heavy tails in the LLM output. We attribute the performance difference to downstream policy sensitivity (e.g., collision avoidance) rather than intermediate target regression quality. We next investigate which specific components of $\mathcal{S}$ contribute most to this regression, and whether the information typically omitted in colloquial language is indeed dispensable.

\subsection{Impact of Communication Components}
\label{subsec:comm_components}

To determine which of the five cues in $\mathcal{S}$ (Sec.~\ref{subsec:comm_message_structure}) drive performance, we ablate each component by masking it at test time (Table~\ref{tab:rebuttal_ablation_components}).

\begin{table}[!ht]
\centering
\scriptsize
\setlength{\tabcolsep}{2pt}
\caption{Ablation of $\mathcal{S}$ components.}
\begin{tabular*}{\linewidth}{@{\extracolsep{\fill}}lccccccc@{}}
\hline
Ablation           & S↑    & S\textsubscript{steps}↓ & SPS↑  & F↑    & CR↓   & CR\textsubscript{T}↓ & ES↑   \\
\midrule
\modelname         & 0.78 & 572      & 0.38  & 0.13 & 0.51 & 0.23    & 0.24 \\
\midrule
$-x_h$              & 0.73 & 595                      & 0.37  & 0.11  & 0.58  & 0.25                 & 0.19  \\
$-x_t$              & 0.77 & 528                      & 0.40  & 0.12  & 0.61  & 0.27                 & 0.19  \\
$-\mathbf{x}_l$     & 0.77 & 555                      & 0.38  & 0.12 & 0.59  & 0.23                 & 0.20  \\
$-\mathbf{x}_d$     & 0.76 & 556                      & 0.37  & 0.12  & 0.55  & 0.24                 & 0.22  \\
$-\mathbf{x}_p$     & 0.78 & 527                      & 0.41  & 0.12  & 0.59  & 0.25                 & 0.18  \\
$-\mathbf{x}_p^{20}$& 0.78 & 517                      & 0.41  & 0.13  & 0.60  & 0.27                 & 0.21  \\
$-\mathbf{x}_p^{50}$& 0.76 & 535                      & 0.40  & 0.13  & 0.58  & 0.25                 & 0.22  \\

\hline
\end{tabular*}
\label{tab:rebuttal_ablation_components}
\end{table}

\modelname~is broadly robust to partial information. The largest drop comes from removing $x_h$ (ES: $0.24{\to}0.19$), which gates whether encounter-related cues are available. Masking $x_t$ or the full speaker trajectory $\mathbf{x}_p$ also reduces ES ($0.19$ and $0.18$), suggesting that temporal recency and the bystander's path help the policy decide how actionable a cue is. This matches the natural-language setting: humans often express these cues approximately, e.g., ``a few seconds ago'' or ``I came from behind'', rather than with precise numerical values.

\subsection{Scalability to Multiple Human Agents}
\label{subsec:scalability}

We evaluate all models with three humanoid agents (Table~\ref{tab:ablation_table_three_humans}). Because domestic scenes in Habitat 3.0c are physically compact, three agents at full speed saturate the walkable area, triggering frequent unavoidable collisions. We additionally halve each humanoid's velocity to emulate larger spaces where agents have room for safer decisions, effectively scaling up the navigable area.

\begin{table}[t]
\centering
\scriptsize
\setlength{\tabcolsep}{3pt}
\renewcommand{\arraystretch}{1.2}
\caption{Three-human evaluation at full and half speed.}
\label{tab:ablation_table_three_humans}
\begin{tabular}{@{} l c c c c c c c c c c @{}}
\hline
Model & Int. & \#H & Speed & S & \(S_{\text{steps}}\) & SPS & F & CR & \(\mathrm{CR}_T\) & ES \\
\hline
DDPPO & - & 3 & 1 & 0.61 & 730 & 0.34 & 0.07 & 0.80 & 0.27 & 0.07 \\
SDA & - & 3 & 1 & 0.62 & 708 & 0.35 & 0.07 & 0.81 & \(\mathbf{0.24}\) & 0.08 \\
COMM & \checkmark & 3 & 1 & \(\mathbf{0.67}\) & 649 & 0.35 & \(\mathbf{0.09}\) & \(\mathbf{0.73}\) & 0.25 & \(\mathbf{0.12}\) \\
$\text{COMM}_{\mathcal{L}}$ & \checkmark & 3 & 1 & \(\mathbf{0.67}\) & \(\mathbf{646}\) & \(\mathbf{0.36}\) & 0.08 & 0.77 & 0.26 & 0.11 \\
\hline
DDPPO & - & 3 & ½ & 0.58 & 799 & 0.37 & 0.10 & 0.49 & 0.15 & 0.21 \\
SDA & - & 3 & ½ & 0.61 & 760 & 0.42 & 0.11 & 0.53 & 0.14 & 0.24 \\
COMM & \checkmark & 3 & ½ & \textbf{0.67} & \textbf{711} & \textbf{0.45} & \textbf{0.13} & \textbf{0.42} & \textbf{0.11} & \textbf{0.29} \\
\hline
\end{tabular}
\end{table}

At full speed, both \modelname~variants raise $S$ to $0.67$ (vs.\ $0.62$/$0.61$ for SDA/DDPPO), reduce $S_{\text{steps}}$ to $649$/$646$ (from $708$/$730$), and attain the highest ES ($0.12/0.11$). At half speed the gap widens: \modelname~achieves ES of $0.29$ vs.\ $0.21$/$0.24$ for DDPPO/SDA, with markedly lower CR ($0.42$ vs.\ $0.49$/$0.53$). Communication-driven guidance thus scales to denser settings and becomes increasingly advantageous as the environment permits safer navigation.

\newline
\noindent\textbf{Limitations:}
Consistent with the most recent works in Social Navigation (\cite{Puig23}, \cite{Scofano24}), our evaluation is conducted in simulation with noiseless sensing and perfect egocentric grounding. Deploying a policy trained in Habitat 3.0c onto a physical robot requires integrating practical modules for person segmentation and identification, components that are assumed to be oracle-level in simulation.

We focus on single-turn interaction to isolate the impact of social grounding without complex dialogue management. While CommNav establishes the baseline for mapless query-based navigation, extending this framework to multi-turn clarification cycles is a natural progression for future research.

Finally, while socially aware navigation holds promise for supporting vulnerable users, it also raises ethical concerns, particularly regarding surveillance. In our experiments, interactions are limited to consenting agents; however, real-world deployment will require stronger safeguards, including explicit consent, privacy-preserving data handling, and clear policy constraints on communication.

\section{Conclusions}

In this work, we introduce Communication-enabled Social Navigation (CommNav), the first framework to explicitly integrate human-robot communication into embodied navigation, along with its supporting simulator, Habitat 3.0c. At the core of our approach is the COMM module, which effectively exploits sparse communicative cues to guide navigation policies.

Our investigation focused not just on \textit{if} communication helped, but on \textit{how} its modality impacts performance. We compared a policy using perfect, structured data (\modelname) with one trained on LLM-generated language (\(\text{\modelname}_{\mathcal{L}}\)). Our key finding emerged when we evaluated this language policy using colloquial instructions from a human study: the policy was exceptionally robust to simple, imprecise human language, achieving episode success statistically indistinguishable from that of the model using structured data.

This result establishes that communication is a key mechanism for scaling navigation to complex, crowded settings. Crucially, it demonstrates that there is a direct and robust path to ground these policies in the natural, colloquial, and imperfect language of human interaction in the real world.


\section*{APPENDIX}

\subsection{Experimental Setup}
Following state-of-the-art social navigation works~\cite{Puig23, Scofano24}, we train on Habitat 3.0c with 24 parallel environments and evaluate on 12 held-out environments. Training runs for 200M steps on 4$\times$A100 GPUs over six days; evaluation uses a single A100 GPU for four hours.

\subsection{\taskname}
At every simulation step, human agents register the information they need to communicate to the robot as follows: $S_{target} = (x_h, x_t, \mathbf{x}_l, \mathbf{x}_d) \in \mathbb{R}^8,$ $S_{agent} = (\mathbf{x}_p) \in \mathbb{R}^{(100,3)}.$

If the agents meet again, all encounter-related variables are overwritten. However, the non-goal agent may inform the robot that they encountered the target agent over $n=100$ steps earlier; in this case, $\mathbf{x}_p$ would no longer contain a valid path to the encounter location. During simulation, the distance $d_p$ covered by a humanoid agent with linear speed $v = 10.0$---inherited from Habitat 3.0~\cite{Puig23}---between two steps averages around 30\,cm, making the communicated trajectory $|\mathbf{x}_p| = n \times d_p$ centimeters long.

\subsection{\modelname~Module and ablation}\label{sec:refer_module_implementation}\label{abl:\modelname}
The TCN Encoder consists of 1D-Convolutional layers with a hidden dimension of $H=64$, dropout rate $p_d=0.2$, and ReLU activation. The Transformer Encoder uses an input dimension of $H$, with $3$ attention heads and $2$ layers, while the ST-MLP features a hidden and output dimension of $H$.

Table~\ref{tab:ablation_table_refer_time} investigates whether the performance improvements of \modelname~result solely from extended training or from its specialized design.




\begin{table}[b]
\centering
\scriptsize
\setlength{\tabcolsep}{3pt}
\renewcommand{\arraystretch}{1.1}
\caption{Ablation on longer training.}
\begin{tabular}{@{} l c c c c c c c c c @{}}
\hline
Model & Int. & Steps & S & $S_{\text{steps}}$ & SPS & F & CR & $\mathrm{CR}_T$ & ES \\
\hline
DDPPO~\cite{Puig23} & - & 200M & 0.71 & 618 & 0.40 & 0.10 & 0.64 & 0.30 & 0.14 \\
DDPPO~\cite{Puig23} & - & 270M & 0.72 & 573 & \textbf{0.42} & 0.10 & 0.72 & 0.24 & 0.16 \\
\hline
DDPPO~\cite{Puig23} & \checkmark & 200M & 0.70 & 608 & 0.40 & 0.10 & 0.68 & 0.30 & 0.14 \\
DDPPO~\cite{Puig23} & \checkmark & 270M & 0.71 & 618 & 0.38 & 0.11 & 0.64 & 0.28 & 0.16 \\
\hline
\modelname\ \textit{(Train)} & \checkmark & 200M & 0.67 & 652 & 0.38 & 0.10 & 0.71 & 0.25 & 0.18 \\
\modelname\ \textit{(Train)} & \checkmark & 270M & 0.64 & 659 & 0.39 & 0.10 & 0.65 & 0.18 & 0.18 \\
\hline
\modelname\ \textit{(Fine-tune)} & \checkmark & 200M+70M & \textbf{0.78} & \textbf{572} & 0.38 & \textbf{0.13} & \textbf{0.51} & \textbf{0.23} & \textbf{0.24} \\
\hline
\end{tabular}
\label{tab:ablation_table_refer_time}
\end{table}

\noindent
We compare DDPPO, DDPPO with Interaction, COMM (\textit{Train}) (training DDPPO with COMM enabled from step 0), and COMM (\textit{Fine-tune}) (fine-tuning DDPPO with Interaction using the frozen COMM module).

For DDPPO without interaction, extending training from 200M to 270M steps yields only marginal improvements (ES: 0.14$\to$0.16). A similar plateau occurs with interaction enabled. In contrast, COMM (\textit{Fine-tune}) attains substantially higher performance (\textit{S} of 0.78, \textit{CR} of 0.51, \textit{ES} of 0.24), while COMM (\textit{Train}) models achieve only moderate scores (\textit{ES} of 0.18). These results confirm that improvements stem from the communication module's design, not prolonged training. Moreover, fine-tuning reduces training time by approximately six days on a 4$\times$A100 setup.

\section*{ACKNOWLEDGMENT}
We acknowledge partial financial support from Panasonic, and the Sapienza grant RG123188B3EF6A80 (CENTS). We acknowledge CINECA for computational resources and support.

\bibliographystyle{IEEEtran}
\bibliography{IEEEabrv,IEEEfull}

\end{document}